\def\year{2019}\relax
\documentclass[letterpaper]{article} 
\usepackage{aaai19}  
\usepackage{times}  
\usepackage{helvet}  
\usepackage{courier}  
\usepackage{url}  
\usepackage{graphicx}  

\usepackage{amsmath}
\usepackage[utf8]{inputenc}
\usepackage{soul}
\newcommand\round[1]{\left[#1\right]}

\begin{document}
%
\title{Non-Autoregressive Machine Translation with Auxiliary Regularization}
\author{
  $^1$Yiren Wang\thanks{The work was done when the author was an intern at Microsoft Research Asia.}, $^2$Fei Tian, $^3$Di He, $^2$Tao Qin, $^1$ChengXiang Zhai, $^2$Tie-Yan Liu\\
  $^1$University of Illinois at Urbana-Champaign, Urbana, IL, USA\\
  $^2$Microsoft Research, Beijing, China\\
  $^3$Key Laboratory of Machine Perception, MOE, School of EECS, Peking University, Beijing, China\\
  $^1$\texttt{\{yiren,czhai\}@illinois.edu}\\
  $^2$\texttt{\{fetia, taoqin, tie-yan.liu\}@microsoft.com}\\
  $^3$\texttt{di\_he@pku.edu.cn}\\
}

\maketitle
\begin{abstract}
As a new neural machine translation approach, Non-Autoregressive machine Translation (NAT) has attracted attention recently due to its high efficiency in inference. However, the high efficiency has come at the cost of not capturing the sequential dependency on the target side of translation, which causes NAT to suffer from two kinds of translation errors: 1) repeated translations (due to indistinguishable adjacent decoder hidden states), and 2) incomplete translations (due to incomplete transfer of source side information via the decoder hidden states).  
In this paper, we propose to address these two problems by improving the quality of decoder hidden representations via two auxiliary regularization terms in the training process of an NAT model. First, to make the hidden states more distinguishable, we regularize the similarity between consecutive hidden states based on the corresponding target tokens. Second, to force the hidden states to contain all the information in the source sentence, we leverage the dual nature of translation tasks (e.g., English to German and German to English) and minimize a backward reconstruction error to ensure that the hidden states of the NAT decoder are able to recover the source side sentence. Extensive experiments conducted on several benchmark datasets show that both regularization strategies are effective and can alleviate the issues of repeated translations and incomplete translations in NAT models. The accuracy of NAT models is therefore improved significantly over the state-of-the-art NAT models with even better efficiency for inference.  
\end{abstract}

\section{Introduction}
Neural Machine Translation (NMT) based on deep neural networks has gained rapid progress over recent years~\cite{s2s_emnlp,nmt_attention,GNMT,transformer,humanNMT}. NMT systems are typically implemented in an encoder-decoder framework, in which the encoder network feeds the representations of source side sentence $x$ into the decoder network to generate the tokens in target sentence $y$. The decoder typically works in an auto-regressive manner: the generation of the $t$-th token $y_t$ follows a conditional distribution $P(y_t|x, y_{<t})$, where $y_{<t}$ represents all the generated tokens before $y_t$. Therefore, during the inference process, we have to sequentially translate each target-side word one by one, which substantially hampers the inference efficiency of NMT systems. 

To alleviate the latency of inference brought by auto-regressive decoding, recently the community has turned to Non-Autoregressive Translation (NAT) systems~\cite{NAT,cho_nat,NAT_google}. A {\em basic NAT model} has the same encoder-decoder architecture as the autoregressive translation (AT) model, except that the sequential dependency within the target side sentence is omitted. In this way, all the tokens can be generated in parallel, and the inference speed is thus significantly boosted. However, it comes at the cost that the translation quality is largely sacrificed since the intrinsic dependency within the natural language sentence is abandoned. To mitigate such performance degradation, previous work has tried different ways to insert intermediate discrete variables to the basic NAT model, so as to incorporate some light-weighted sequential information into the non-autoregressive decoder. The discrete variables include the autoregressively generated latent variables~\cite{NAT_google}, and the fertility information brought by a third-party word alignment model~\cite{NAT}. However, leveraging such discrete variables not only brings additional difficulty for optimization, but also slows down the translation by introducing extra computational cost for producing such discrete variables. 

In this paper, we propose a new solution to the problem that does not rely on any discrete variables and makes little revision to the basic NAT model, thus retaining most of the benefit of an NAT model. Our approach was motivated by the following result we obtained from carefully analyzing the key issues existed in the basic NAT model. We empirically observed that the two types of translation errors frequently made by the basic NAT model are: 1) {\em repeated translation}, where the same token is generated repeatedly at consecutive time steps; 2) {\em incomplete translation}, where the semantics of several tokens/sentence pieces from the source sentence are not adequately translated. Both issues suggest that the decoder hidden states in NAT model, i.e., the hidden states output in the topmost layer of the decoder, are of low quality: the repeated translation shows that two adjacent hidden states are {\em indistinguishable}, leading to the same tokens decoded out, while the incomplete translation reflects that the hidden states in the decoder are {\em incomplete} in representing source side information. 

Such a poor quality of decoder hidden states is in fact a direct consequence of the non-autoregressive nature of the model: at each time step, the hidden states have no access to their prior decoded states, making them in a `chaotic' state being unaware of what has and has not been translated. Therefore, it is difficult for the neural network to learn good hidden representations all by itself. Thus, in order to improve the quality of the decoder representations, we must go beyond the pure non-autoregressive models. The challenge, though, is how to improve the quality of decoder representations to address the two problems identified above while still keeping the benefit of efficient inference of the NAT models.  

We propose to address this challenge by directly regularizing the learning of the decoder representations using two auxiliary regularization terms for model training. First, to overcome the problem of repeated translation, we propose to force the similarity of two neighboring hidden state vectors to be well aligned with the similarity of the embedding vectors representing the two corresponding target tokens they aim to predict. We call this regularization strategy {\em similarity regularization}. Second, to overcome the problem of incomplete translation, inspired by the dual nature of machine translation task~\cite{dual}, we propose to impose that the hidden representations should be able to reconstruct the source side sentence, which can be achieved by putting an auto-regressive backward translation model on top of the decoder; the backward translation model would ``demand" the decoder hidden representations of NAT to contain enough information about the source.  We call this regularization strategy {\em reconstruction regularization}. Both regularization terms only appear in the training process and have no effect during inference, thus bringing no additional computational/time cost to inference and allowing us to retain the major benefit of an NAT model. Meanwhile, the {\em direct} regularization on the hidden states effectively improves their representation. In contrast, the existing approaches would need either a third-party word alignment module~\cite{NAT} (hindering end-to-end learning), or a special component handling the discrete variables (e.g., several softmax operators in~\cite{cho_nat}) which brings additional latency for decoding. 

We evaluate the proposed regularization strategies by conducting extensive experiments on several benchmark machine translation datasets. The experiment results show that 
both regularization strategies are effective and they can alleviate the issues of repeated translations and incomplete translations in NAT models, leading to improved NAT models that can improve accuracy substantially over the state-of-the-art NAT models without sacrificing efficiency. We set a new state-of-the-art performance of non-autoregressive translation models on the WMT14 datasets, with $24.61$ BLEU in En-De and $28.90$ in De-En.

\section{Background}

{\em Neural machine translation}~\cite{nmt_attention} (NMT) has been the widely adopted machine translation approach within both academia and industry. A neural machine translation system specifies a conditional distribution $P(y|x)$ of the likelihood of translating source side sentence $x$ into target sentence $y$. There are typically two parts in NMT: the encoder network and the decoder network. The encoder reads the source side sentence $x=(x_1,\cdots, x_{T_x})$ with $T_x$ tokens, and processes it into context vectors which are then fed into the decoder network. The decoder builds the conditional distribution $P(y_t|y_{<t},x)$ at each decoding time-step $t$, where $y_{<t}$ represents the set of generated tokens before time-step $t$. The final distribution $P(y|x)$ is then in the factorized form $P(y|x)=\prod_{t=1}^{T_y}P(y_t|y_{<t},x)$, with $y$ containing $T_y$ target side tokens $y=(y_1,\cdots, y_{T_y})$. 

The NMT suffers from the high inference latency, which limits its application in the real world scenarios. The main bottleneck comes from its autoregressive nature of the sequence generation: each target side token is generated one by one, which prevents parallelism during inference, and thus the computational power of GPU cannot be fully exploited. Recent research efforts have been resorted to solve the huge latency issue in the decoding process. In the domain of speech synthesis, parallel wavenet~\cite{parallel_wavenet} successfully achieves the parallel sampling based on inverse autoregressive flows~\cite{iaf} and knowledge distillation~\cite{kd}. In NMT, the non-autoregressive neural machine translation (NAT) model has been developed recently~\cite{NAT,cho_nat,NAT_google}. In NAT, all the tokens within one target side sentence are generated in parallel, without any limitation of sequential dependency. The inference speed has thus been significantly boosted (e.g., tens of times faster than the typical autoregressive NMT model) and for the sake of maintaining translation quality, there are several technical innovations in the NAT model design from previous exploration:

\begin{figure*}[t]
\centering
\includegraphics[width=.95\linewidth]{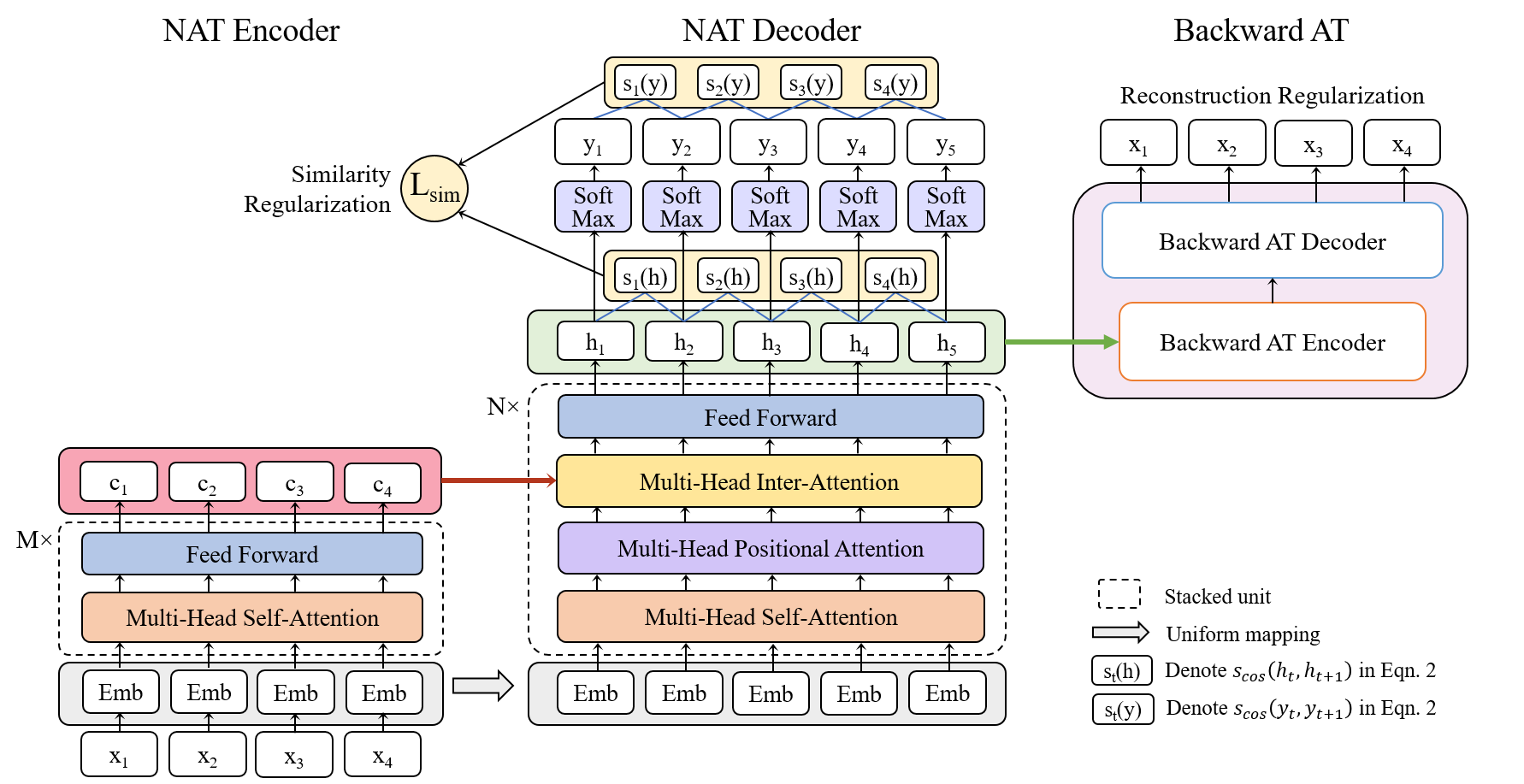}
\caption{The overall architecture of NAT model with auxiliary regularization terms. {\em AT} stands for autoregressive translation. Each sublayer of the encoder/decoder contains a residual connection and layer normalization following \cite{transformer}.}
\label{fig:frame}
\end{figure*}

\begin{itemize}
    \item {\em Sequence level knowledge distillation}. NAT model is typically trained with the help from an autoregressive translation (AT) model as its teacher. The knowledge of the AT model is distilled to the NAT model via the sequence level distillation technique~\cite{sequence_kd} which is essentially using the sampled translation from the teacher model, out from the source side sentences, as the bilingual training data. It has been reported previously~\cite{NAT}, and is also consistent with our empirical studies that training NAT with such distilled data performs much better than the original ground truth data, or the mixture of ground truth and distilled translations. There has been no precise theoretical justification, but an intuitive explanation is that the NAT model suffers from the `multimodality' problem~\cite{NAT} -- there might be multiple valid translations for the same source word/phrase/sentence. The distilled data via the teacher model eschew such problem since the output of a neural network is less noisy and more `deterministic'. 
    
    \item {\em Model architecture modification}. The NAT model is similar in the framework of encoder and decoder but has several differences from the autoregressive model, including (a) the causal mask of decoder, which prevents the earlier decoding steps from accessing later information, is removed; (b) positional attention is leveraged in the decoder to enhance position information, where the positional embeddings \cite{transformer} are used as query and key, and hidden representations from the previous layers are used as value \cite{NAT}.
    
    \item {\em Discrete variables to aid NAT model training and inference}. There are intermediate discrete variables in previous NAT models, which aim to make up the loss of sequential information in the non-autoregressive decoding. The representative examples include the fertility value to indicate the number of copying each source side token~\cite{NAT}, and the discrete latent variables autoregressively generated ~\cite{NAT_google}. Additional difficulty arises due to the discreteness, and thus specially designed optimization processes are necessary such as variational method~\cite{NAT,cho_nat} and vector quantization~\cite{NAT_google}.
\end{itemize}

\section{Model}
We introduce the details of our proposed approaches to non-autoregressive neural machine translation in this section. The overall model architecture is depicted in Fig.~\ref{fig:frame}. 

We use the basic NAT model with the same encoder-decoder architecture as the AT model, and follow two of the aforementioned techniques of training NAT models. First, we train our NAT models based on the sequence-level knowledge distillation. Second, we obey the aforementioned model variants specially designed for NAT, such as the positional attention and removing causal mask in the decoder~\cite{NAT}. What makes our solution unique is that we replace the hard-to-optimize discrete variables in prior works with two simple regularization terms. As a pre-requisite, since there are no discrete variables available to indicate some sequential information, we need new mechanisms to {\em predict the target length during inference} and {\em generate inputs for the decoder}, which are necessary for the NAT model since the information fed into the decoder (i.e., the word embeddings of previous target side tokens in autoregressive model) is unknown due to parallel decoding. 

{\em Target length prediction during inference}. Instead of estimating target length $T_y$ with fertility values~\cite{NAT}, or with a separate model $p(T_y|x)$~\cite{cho_nat}, we use a simple yet effective method as~\cite{hint-based} that leverages source length to determine target length. During inference, the target side length $T_y$ is predicted as $T_y = T_x + \Delta T$, where $T_x$ is the length of source sentence, $\Delta T$ is a constant bias term that can be set according to the overall length statistics of the training data. 
    
{\em Generate decoder input with uniform mapping}. Different from other works relying on the fertility~\cite{NAT}, we adopt a simple way of uniformly mapping the source side word embedding as the inputs to the decoder. Specifically, given source sentence $x = \{x_1, x_2, ..., x_{T_x}\}$ and target length $T_y$, we uniformly map the embeddings (denoted as $\mathcal{E}(\cdot)$) of the source tokens to generate decoder input with:  
    \begin{equation}
    \label{eqn:decoder-input}
        z_k = \mathcal{E}(x_i), \ i = \round{\frac{T_x}{T_y}t}, \ t=1,2,...,T_y.
    \end{equation}
    
We use this variant of the basic NAT model as our backbone model (denoted as ``{\em NAT-BASE}"). We now turn to analyze the two commonly observed translation issues of NAT-BASE and introduce the corresponding solutions to tackle them.

\subsection{Repeated Translation \& Similarity Regularization}

The issue of {\em repeated translation} refers to the same word token being successively generated for multiple times, e.g., the ``mobility mobility , , s'' in the second example of Table~\ref{tbl:case}. Such an issue is quite common for NAT models. On IWSLT14 De-En dataset, we empirically find that more than half of NAT translations contain such repeated tokens. Furthermore, simply de-duplicating such repeated tokens brings limited gain (shown in Table~\ref{tbl:ablation}). This is consistent with our observation that very likely there will be additional translation errors close to the repeated tokens, e.g., ``caniichospital rates'' in the same example. Apparently, it reflects that the decoder hidden states at consecutive time steps are largely similar, if the same tokens are decoded out from them. 

One natural way to tackle these indistinguishable representations is to impose a similarity regularization term to force the two hidden states to be far away. Here the similarity measure to be minimize could be, for example, the cosine similarity $s_{cos}(h, h')=\frac{h h'^T}{||h||\cdot||h'||}$ and L2 similarity $s_{l2}(h,h')=-||h-h'||_2^2$. Acting in this way is simple and straightforward, but it is also apparently problematic: the degree of closeness for each pair of adjacent hidden states $(h_t, h_{t+1})$ varies with data. Universally imposing the same similarity regularization term for all data $(x,y)$ and all time-step $t$ does not accommodate such flexibility. For example, consider the target side phrase $(y_t,y_{t+1}, y_{t+2})$=`hollywood movie 2019', the hidden state $h_{t+1}$ should be closer to $h_{t}$ than with $h_{t+2}$, since the token `movie' is semantically more related with `hollywood' than `2019'. Then adding the same constraint on $(h_t, h_{t+1})$ and $(h_{t+1}, h_{t+2})$ is sub-optimal. 

We therefore adopt a more flexible way to specify the degree of similarity regularization on two adjacent hidden states. Our intuition is consistent with the example above: for $(h_t, h_{t+1})$, if the target tokens $(y_t, y_{t+1})$ which are decoded from them are semantically far apart, then the similarity between $h_t$ and $h_{t+1}$ should be correspondingly smaller, and vice versa. Here the semantic similarity between two word tokens $y_t$ and $y_{t+1}$ is calculated with their word embeddings~\cite{word2vec}. The regularization term between $h_t$ and $h_{t+1}$ then goes as:

\begin{equation}
    L_{sim,t}=1+s_{cos}(h_t,h_{t+1})\cdot (1-s_{cos}(y_t, y_{t+1})),
\end{equation}
where with a little abuse of notations, $y_t$ also represents the target side word embedding in the NMT model for word $y_t$. The regularization term basically says, if two words $y_t$ and $y_{t+1}$ are dissimilar ($s_{cos}(y_t, y_{t+1})$ is smaller), then the similarity regularization (i.e., $s_{cos}(h_t, h_{t+1})$) should have larger effect, and vice versa. Here the constant $1$ is for keeping the loss values non-negative. Note here we do not incur gradients on the word embeddings $y_t$ and $y_{t+1}$ via Eqn.~\ref{eqn:dis_reg}, for fear of bad effects brought by unqualified hidden states to the training of word embeddings. The overall loss on the sample $(x,y)$ then is: 

\begin{equation}
\label{eqn:dis_reg}
L_{sim}=\sum_t^{T_y}L_{sim,t}.    
\end{equation}

\subsection{Incomplete Translation \& Reconstruction Regularization}

The {\em incomplete translation} means that several tokens/phrases in the source sentence are neglected by the translation model. Previously the autoregressive translation model was also observed to suffer from incomplete translation~\cite{NMTCoverage}. However, we empirically find that with the more powerful backbone model, such as the deep Transformer, incomplete translation is not common for autoregressive translation, as is shown by the outputs of the teacher model in Table~\ref{tbl:case} (labeled by `AT'). However, NAT suffers from incomplete translation significantly, such as in the same table, the source side phrase `und manches nicht' in the third example is totally missed by the NAT model. We further take a close look at $50$ randomly non-autoregressive translation results sampled from IWSLT14 De-En dataset, and find that more than $30\%$ of them leave something behind on the source side. 

The reconstruction regularization targets at solving the incomplete translation problem. Concretely speaking, we borrow the spirit of dual learning~\cite{dual} by bringing in an additional backward translation model $P(x|y)$ which we further denote as $\overleftarrow{P}$ for the ease of statement. We couple the two models (i.e., the original NAT model $P$ translating $x$ to $y$ and the backward translation model $\overleftarrow{P}$ translating $y$ to $x$) together by feeding the hidden states output by the decoder of $P$ (i.e., $h=\{h_1,\cdots, h_{T_y}\}$) to $\overleftarrow{P}$, and asking $\overleftarrow{P}$ to recover the source sentence from $h$. Here $h$ simply acts as the input embeddings for the encoder of $\overleftarrow{P}$. The reconstruction loss for a training data $(x,y)$ then goes as:

\begin{equation}
\label{eqn:res_reg}
    L_{rec}=-\sum_{t=1}^{T_x}\log \overleftarrow{P}(x_t|h, x_{<t}).
\end{equation} 

We set the backward translation model $\overleftarrow{P}$ as the autoregressive model since autoregressive model provides more accurate information. $\overleftarrow{P}$ will not be used in the inference process of $P$. One can easily see $\overleftarrow{P}$ forces $h$ to be more comprehensive in storing the source side information by directly providing $x$ as training objective. 

Here another intuitive view to understand the effect of $\overleftarrow{P}$ is that it acts in a similar way as the discriminator of generative adversarial networks~\cite{GAN}: $\overleftarrow{P}$ decides whether $h$ is good enough by testing whether $x$ can be perfectly recovered from $h$. Then the NAT model $P$ plays the effect of the generator with the hope of reconstructing $x$, and thus fooling $\overleftarrow{P}$. Setting $\overleftarrow{P}$ to be autoregressive makes the `discriminator' not easily fooled, urging the output of `generator', i.e., the hidden state $h$, to be as complete as possible to deliver the information from source side.

There are prior works for enhancing the adequacy of translation results, but devoted to autoregressive NMT models~\cite{NMTContextualGates,NMTCoverage,NMTReconstruction}.  Among all these prior works, our approach is most technically similar with that in~\cite{NMTReconstruction}, with the difference that for reconstructing from $h$ to $x$, we leverage a complete backward translation model based on encoder-decoder framework, while they adopt a simple decoder without encoder. We empirically find that such an encoder cannot be dropped in the scenario of helping NAT models, which we conjecture is due to the better expressiveness brought by the encoder. Another difference is that they furthermore leverage the reconstruction score to assist the inference process via re-ranking different candidates, while we do not perform such a step since it hurts the efficiency of inference.

\begin{table*}[h]
\centering
\begin{tabular}{lcccccc}
\hline
\multicolumn{1}{l|}{Models/Datasets}                                  & \multicolumn{1}{c|}{\begin{tabular}[c]{@{}c@{}}WMT14 \\ En-De\end{tabular}} & \multicolumn{1}{c|}{\begin{tabular}[c]{@{}c@{}}WMT14 \\ De-En\end{tabular}} & \multicolumn{1}{c|}{\begin{tabular}[c]{@{}c@{}}IWSLT14 \\ De-En\end{tabular}} & \multicolumn{1}{c|}{\begin{tabular}[c]{@{}c@{}}IWSLT16 \\ En-De\end{tabular}} & \multicolumn{1}{c|}{Latency}          & Speedup       \\ \hline
{\em Autoregressive Models (AT Teachers)}                          &                                                                             &                                                                             &                                                                               &                                                                               &                                       &               \\ \hline
\multicolumn{1}{l|}{Transformer (NAT-FT)}                 & \multicolumn{1}{c|}{$23.45$}                                                & \multicolumn{1}{c|}{$27.02$}                                                & \multicolumn{1}{c|}{$31.47^\dagger$}                                          & \multicolumn{1}{c|}{$29.70$}                                                  & \multicolumn{1}{c|}{--}               & --            \\
\multicolumn{1}{l|}{Transformer (NAT-IR)}             & \multicolumn{1}{c|}{$24.57$}                                                & \multicolumn{1}{c|}{$28.47$}                                                & \multicolumn{1}{c|}{$30.90^\dagger$}                                          & \multicolumn{1}{c|}{$28.98$}                                                  & \multicolumn{1}{c|}{--}               & --            \\
\multicolumn{1}{l|}{Transformer (LT)}             & \multicolumn{1}{c|}{$27.3$}                                                 & \multicolumn{1}{c|}{/}                                                      & \multicolumn{1}{c|}{/}                                                        & \multicolumn{1}{c|}{/}                                                        & \multicolumn{1}{c|}{--}               & --            \\
\multicolumn{1}{l|}{\textbf{Transformer (NAT-REG)}}                   & \multicolumn{1}{c|}{$27.3$}                                                 & \multicolumn{1}{c|}{$31.29$}                                                & \multicolumn{1}{c|}{$33.52$}                                                  & \multicolumn{1}{c|}{$28.35$}                                                  & \multicolumn{1}{c|}{$607$ ms}         & $1.00 \times$ \\
\multicolumn{1}{l|}{\textbf{Transformer (NAT-REG, Weak Teacher)}} & \multicolumn{1}{c|}{$24.50$}                                                & \multicolumn{1}{c|}{$28.76$}                                                & \multicolumn{1}{c|}{/}                                                        & \multicolumn{1}{c|}{/}                                                        & \multicolumn{1}{c|}{--}               & --            \\ \hline
{\em Non-Autoregressive Models}                                    &                                                                             &                                                                             &                                                                               &                                                                               &                                       &               \\ \hline
\multicolumn{1}{l|}{NAT-FT (no NPD)}                                  & \multicolumn{1}{c|}{$17.69$}                                                & \multicolumn{1}{c|}{$21.47$}                                                & \multicolumn{1}{c|}{$20.32^\dagger$}                                          & \multicolumn{1}{c|}{$26.52$}                                                  & \multicolumn{1}{c|}{$39$ ms}          & $15.6 \times$ \\
\multicolumn{1}{l|}{NAT-FT (NPD rescoring $10$)}                      & \multicolumn{1}{c|}{$18.66$}                                                & \multicolumn{1}{c|}{$22.41$}                                                & \multicolumn{1}{c|}{$21.39^\dagger$}                                          & \multicolumn{1}{c|}{$27.44$}                                                  & \multicolumn{1}{c|}{$79$ ms}          & $7.68 \times$ \\
\multicolumn{1}{l|}{NAT-FT (NPD rescoring $100$)}                     & \multicolumn{1}{c|}{$19.17$}                                                & \multicolumn{1}{c|}{$23.20$}                                                & \multicolumn{1}{c|}{$24.21^\dagger$}                                          & \multicolumn{1}{c|}{$28.16$}                                                  & \multicolumn{1}{c|}{$257$ ms}         & $2.36 \times$ \\
\multicolumn{1}{l|}{NAT-IR ($1$ refinement)}                          & \multicolumn{1}{c|}{$13.91$}                                                & \multicolumn{1}{c|}{16.77}                                                  & \multicolumn{1}{c|}{$21.86^\dagger$}                                          & \multicolumn{1}{c|}{$22.20$}                                                  & \multicolumn{1}{c|}{$68^\dagger$ ms}  & $8.9 \times$  \\
\multicolumn{1}{l|}{NAT-IR ($10$ refinements)}                        & \multicolumn{1}{c|}{$21.61$}                                                & \multicolumn{1}{c|}{$25.48$}                                                & \multicolumn{1}{c|}{$23.94^\dagger$}                                          & \multicolumn{1}{c|}{$27.11$}                                                  & \multicolumn{1}{c|}{$404^\dagger$ ms} & $1.5 \times$  \\
\multicolumn{1}{l|}{NAT-IR (adaptive refinements)}                    & \multicolumn{1}{c|}{$21.54$}                                                & \multicolumn{1}{c|}{$25.43$}                                                & \multicolumn{1}{c|}{$24.63^\dagger$}                                          & \multicolumn{1}{c|}{$27.01$}                                                  & \multicolumn{1}{c|}{$320^\dagger$ ms} & $1.9 \times$  \\
\multicolumn{1}{l|}{LT (no rescoring)}                                & \multicolumn{1}{c|}{$19.8$}                                                 & \multicolumn{1}{c|}{/}                                                      & \multicolumn{1}{c|}{/}                                                        & \multicolumn{1}{c|}{/}                                                        & \multicolumn{1}{c|}{$105$ ms}         & $5.78 \times$ \\
\multicolumn{1}{l|}{LT (rescoring $10$)}                              & \multicolumn{1}{c|}{$21.0$}                                                 & \multicolumn{1}{c|}{/}                                                      & \multicolumn{1}{c|}{/}                                                        & \multicolumn{1}{c|}{/}                                                        & \multicolumn{1}{c|}{/}                & /             \\
\multicolumn{1}{l|}{LT (rescoring $100$)}                             & \multicolumn{1}{c|}{$22.5$}                                                 & \multicolumn{1}{c|}{/}                                                      & \multicolumn{1}{c|}{/}                                                        & \multicolumn{1}{c|}{/}                                                        & \multicolumn{1}{c|}{/}                & /             \\ \hline
\multicolumn{1}{l|}{\textbf{NAT-REG (no rescoring)}}                  & \multicolumn{1}{c|}{$20.65$}                                                & \multicolumn{1}{c|}{$24.77$}                                                & \multicolumn{1}{c|}{$23.89$}                                                  & \multicolumn{1}{c|}{$23.14$}                                                  & \multicolumn{1}{c|}{$22$ ms}          & $27.6 \times$ \\
\multicolumn{1}{l|}{\textbf{NAT-REG (rescoring $9$)}}                 & \multicolumn{1}{c|}{$24.61$}                                                & \multicolumn{1}{c|}{$28.90$}                                                & \multicolumn{1}{c|}{$28.04$}                                                  & \multicolumn{1}{c|}{$27.02$}                                                  & \multicolumn{1}{c|}{$40$ ms}          & $15.1 \times$ \\
\multicolumn{1}{l|}{\textbf{NAT-REG (WT, no rescoring)}}              & \multicolumn{1}{c|}{$19.15$}                                                & \multicolumn{1}{c|}{$23.20$}                                                & \multicolumn{1}{c|}{/}                                                        & \multicolumn{1}{c|}{/}                                                        & \multicolumn{1}{c|}{--}               & --            \\
\multicolumn{1}{l|}{\textbf{NAT-REG (WT, rescoring $9$)}}             & \multicolumn{1}{c|}{$22.80$}                                                & \multicolumn{1}{c|}{$27.12$}                                                & \multicolumn{1}{c|}{/}                                                        & \multicolumn{1}{c|}{/}                                                        & \multicolumn{1}{c|}{--}               & --            \\ \hline
\end{tabular}
\caption{The test set performances of AT and NAT models in BLEU score. {\em NAT-FT}, {\em NAT-IR} and {\em LT} denotes the baseline method in~\cite{NAT},~\cite{cho_nat} and~\cite{NAT_google} respectively. {\em NAT-REG} is our proposed NAT with two regularization terms and {\em Transformer (NAT-REG)} is correspondingly our AT teacher model. `Weak Teacher' or `WT' refers to the NAT trained with weakened teacher comparable to prior works. All AT models are decoded with a beam size of $4$. `$\dagger$' denotes baselines from our reproduction. `--' denotes same numbers as above/below.
}
\label{tbl:results}
\end{table*}

\subsection{Joint Training}

In the training process, all the two regularization terms are combined with the commonly used cross-entropy loss $L_{ce}$. For a training data $(x,y)$ in which $y$ is the sampled translation of $x$ from the teacher model, the total loss is:

\begin{align}
\label{eqn:total_loss}
L & = L_{ce} +\alpha L_{sim} + \beta L_{rec} \\
L_{ce}&=\sum_{t=1}^{T_y}\log P(y_t|x)
\end{align}
where $\alpha$ and $\beta$ are the trade-off parameters, $L_{sim}$ and $L_{rec}$ respectively denotes the similarity regularization in Eqn.\ref{eqn:dis_reg} and the reconstruction regularization in Eqn.~\ref{eqn:res_reg}. We denote our model with the two regularization terms as {\em ``NAT-REG''}

\section{Evaluation}

We use multiple benchmark datasets to evaluate the effectiveness of the proposed approach. We compare the proposed approach with multiple baseline approaches that represent the state-of-the-art in terms of both translation accuracy and efficiency, and analyze the effectiveness of the two regularization strategies. As we will show, the strategies are effective and can lead to substantial improvements of translation accuracy with even better efficiency. 

\subsection{Experiment Design}

\subsubsection{Datasets} 
We use several widely adopted benchmark datasets to evaluate the effectiveness of our proposed method: the IWSLT14 German to English translation (IWSLT14 De-En), the IWSLT16 English to German (IWSLT16 En-De), and the WMT14 English to German (WMT14 En-De) and German to English (WMT14 De-En) which share the same dataset. For IWSLT16 En-De, we follow the dataset construction in \cite{NAT,cho_nat} with roughly $195k/1k/1k$ parallel sentences as training/dev/test sets respectively. We use the fairseq recipes\footnote{\url{https://github.com/pytorch/fairseq}} for preprocessing data and creating dataset split for the rest datasets. The IWSLT14 De-En contains roughly $153k/7k/7k$ parallel sentences as training/dev/test sets. The WMT14 En-De/De-En dataset is much larger with $4.5M$ parallel training pairs. We use Newstest2014 as test set and Newstest2013 as dev set. We use byte pair encoding (BPE)~\cite{BPE} to segment word tokens into subword units, forming a $32k$-subword vocabulary shared by source and target languages for each dataset.

\subsubsection{Baselines}
We include three latest NAT works as our baselines, the NAT with fertility (NAT-FT)~\cite{NAT}, the NAT with iterative refinement (NAT-IR)~\cite{cho_nat} and the NAT with discrete latent variables (LT)~\cite{NAT_google}. For all our four tasks, we obtain the baseline performance by either directly using the performance figures reported in the previous works if they are available or producing them by using the open source implementation of baseline algorithms on our datasets. 

\subsubsection{Model Settings}

For the teacher model, we follow the same setup with previous NAT works to adopt the Transformer model~\cite{transformer} as teacher model for both sequence-level knowledge distillation and inference re-scoring. We use the same model size and hyperparameters for each NAT model and its respective AT teacher. Here we note that several baseline results are fairly weak probably due to the weak teacher models. For example, \cite{NAT} reports an AT teacher with $23.45$ BLEU for WMT14 En-De task, while the official number of Transformer is $27.3$~\cite{transformer}. For fair comparison, we bring in a weakened AT teacher model with the same model architecture, yet sub-optimal performances similar to the teacher models in previous works. We rerun our algorithms with such weakened teacher models.

For the NAT model, we similarly adopt the Transformer architecture. For WMT datasets, we use the hyperparameter settings of \texttt{base} Transformer model in~\cite{transformer}. For IWSLT14 DE-EN, we use the \texttt{small} Transformer setting with a $5$-layer encoder and $5$-layer decoder (size of hidden states and embeddings is $256$, and the number of attention heads is $4$). For IWSLT16 EN-DE, we use a slightly different version of small settings with $5$ layers from~\cite{NAT}, where size of hidden states and embeddings are $278$, number of attention heads is $2$. For our models with reconstruction regularization, the backward AT models share word embeddings and the same model size with the NAT model. Our models are implemented based on the official TensorFlow implementation of Transformer\footnote{\url{https://github.com/tensorflow/tensor2tensor}.}.

For sequence-level distillation, we set beam size to be $4$. For our model NAT-REG, we determine the trade-off parameters, i.e, $\alpha$ and $\beta$ in Eqn.~\ref{eqn:total_loss} by the BLEU on the IWSLT14 De-En dev set, and use the same values for all other datasets. The optimal values are $\alpha=2$ and $\beta=0.5$. 

\subsection{Training and Inference}

We train all models using Adam following the optimizer settings and learning rate schedule in Transformer\cite{transformer}. We run the training procedure on $8/1$ Nvidia M40 GPUs for WMT and IWSLT datasets respectively. Distillation and inference are run on $1$ GPU. 

For inference, we follow the common practice of noisy parallel decoding~\cite{NAT}, which generates a number of decoding candidates in parallel and selects the best translation via re-scoring using AT teacher. In our scenario, we generate multiple translation candidates by predicting different target lengths $T_y \in [T_x + \Delta T - B, T_x + \Delta T + B]$, which results in $2B+1$ candidates. In our experiments, we set $\Delta T$ to $2$, $-2$, $-1$, $1$ for WMT14 EN-De, De-En and IWSLT14 De-En, IWSLT16 En-De respectively according to the average length distribution in the training set.

We evaluate the model performances with tokenized case-sensitive BLEU\footnote{\url{https://github.com/moses-smt/mosesdecoder/blob/master/scripts/generic/multi-bleu.perl}}~\cite{bleu} for WMT14 datasets and tokenized case-insensitive BLEU for IWSLT14 datasets. Latency is computed as average per sentence decoding time (ms) on the full test set of IWSLT14 DE-EN without minibatching. We test latency on $1$ NVIDIA Tesla P100 to keep in line with previous works \cite{NAT}.

\begin{table*}[]
\centering
\begin{tabular}[t]{r|l}
\hline
Source:    & bei der coalergy sehen wir klimaveränderung als eine ernste gefahr für unser geschäft .                                                                                                                                                                            \\
Reference: & at coalergy we view climate change as a very serious threat to our business .                                                                                                                                                                                      \\
AT:       & in coalergy , we see climate change as a serious threat to our business .                                                                                                                                                                                          \\
NAT-BASE:       & \begin{tabular}[t]{@{}l@{}}in the coalergy , we \&apos;ll see \textbf{climate climate change change} as a most serious danger \\ for our business . \end{tabular}                                                                                                                           \\
NAT-REG:   & at coalergy , we \&apos;re seeing climate change as a serious threat to our business .                                                                                                                                                                                   \\ \hline
Source:    & \begin{tabular}[t]{@{}l@{}}dies ist die großartigste zeit , die es je auf diesem planeten gab , egal , welchen maßstab sie  \\ anlegen :gesundheit , reichtum , mobilität , gelegenheiten , sinkende krankheitsraten .\end{tabular}                                \\
Reference: & \begin{tabular}[t]{@{}l@{}}this is the greatest time there \&apos;s ever been on this planet by any measure that you wish \\ to choose : health , wealth , mobility , opportunity , {\em declining rates of disease} .\end{tabular}                                   \\
AT:        & \begin{tabular}[t]{@{}l@{}}this is the greatest time you \&apos;ve ever had on this planet , no matter what scale you \\ \&apos;re putting : health , wealth , mobility , opportunities , declining disease rates .\end{tabular}                                               \\
NAT-BASE:       & \begin{tabular}[t]{@{}l@{}}this is the most greatest time that ever existed on this planet no matter what scale they \\ \&apos;re \textbf{imsi :  :  ,  , mobility mobility ,  , scaniichospital rates} .\end{tabular}                \\
NAT-REG:   & \begin{tabular}[t]{@{}l@{}}this is the greatest time that we \&apos;ve ever been on this planet no matter what scale  they \\ \&apos;re \textbf{ianition} : health , wealth , mobility , opportunities , {\em declining disease rates} .\end{tabular} \\ \hline
Source:    & und manches davon hat funktioniert und manches nicht .                                                                                                                                                                                                             \\
Reference: & and some of it worked , {\em and some of it didn \&apos;t .}                                                                                                                                                                                                          \\
AT:        & and some of it worked {\em and some of it didn \&apos;t work .}                                                                                                                                                                                                       \\
NAT-BASE:       & and some of it worked .                                                                                                                                                                                                                                            \\
NAT-REG:   & and some of it worked {\em and some not .}                                                                                                                                                                                                            \\ \hline
\end{tabular}
\caption{Translation examples from IWSLT14 De-En task. The AT result is decoded with a beam size of $4$ and NAT results are generated by re-scoring $9$ candidates. We use the italic fonts to indicate the translation pieces where the NAT has the issue of incomplete translation, and bond fonts to indicate the issue of repeated translation.}
\label{tbl:case}
\end{table*}

\subsection{Results}
We report all the results in Table~\ref{tbl:results}, from which we can make the following conclusions: 

\noindent
1. {\bf Our model improves NAT translation quality with a large margin}. On all the benchmark datasets except for IWSLT16 En-De, our NAT-REG achieves the best translation quality. On the WMT datasets, NAT-REG with strong teacher model has achieved new state-of-the-art performances for NAT, with $24.61$ in En-De and $28.90$ in De-En, which even outperform AT teachers in previous works. On the small dataset of IWSLT16 En-De, due to the inferior of our teacher model ($28.35$ vs. $29.70$ of NAT-FT), our performance is slightly weaker than the previous works.
   
\noindent 
2. {\bf The improvements over baseline model are not merely brought by the stronger teacher models.} In both WMT tasks, our model achieves better performances with weaker AT teacher model that is on par with the teacher model in previous works (e.g. $27.12$ of NAT-REG (weak teacher) vs. $25.48$ of NAT-IR and $22.41$ of NAT-FT on WMT14 De-En). Therefore, it is clear that our proposal to replace hard-to-optimize discrete variables with the simple regularization terms, can indeed help obtain better NAT models. 
    
\noindent
3. {\bf Our NAT model achieves significantly better latency without the intermediate discrete variables hampering inference efficiency}. For example, the speedup compared with AT model is $15.1\times$ when re-scoring $9$ candidates, which is comparable to NAT-FT decoding without re-scoring ($15.6\times$). This further verifies that by removing the intermediate discrete variables, simply decoding via the neural networks is beneficial to improve the inference speed. 

\subsection{Analysis}

\subsubsection{Case Study}

We present several translation examples sampled from the IWSLT14 De-En dataset in Table~\ref{tbl:case}, including the source sentence, the target reference (i.e., the ground-truth translation), the translation given by the teacher model ({\em AT}), by the basic NAT with sequence distillation ({\em NAT-BASE}), and by our NAT with the two regularization terms ({\em NAT-REG}). As can be seen, {\em NAT-BASE} suffers severely from the issue of {\em repeated translation} (e.g., the `climate climate change change' in the first example) and {\em incomplete translation} (e.g., incomplete translation of `and some of it didn \&apos;t' in the third example), while with the two auxiliary regularization terms brought in, the two issues are largely alleviated.

\begin{table}[]
\centering
\begin{tabular}{l|l}
\hline
\multicolumn{1}{c|}{Model variants} & \multicolumn{1}{c}{BLEU} \\ \hline
NAT-BASE &  $28.73$ \\ \hline
NAT-BASE + de-duplication &  $29.45$ \\ 
NAT-BASE + universally penalize similarity & $28.32$ \\ \hline
NAT-BASE + similarity regularization  &  $30.02$ \\
NAT-BASE + reconstruction regularization &  $30.21$
 \\ \hline
\begin{tabular}[c]{@{}l@{}}NAT-BASE  + both regularizations\end{tabular} & $30.84$ \\ \hline
\end{tabular}
\caption{Ablation study on IWLST14 De-En dev set. Results are BLEU scores with teacher rescoring $9$ candidates.}
\label{tbl:ablation}
\end{table}

\subsubsection{Ablation Study}
\label{subsubsec:ablation}

To further study the effects brought by different techniques, we show in Table~\ref{tbl:ablation} the translation performance of different NAT model variants for the IWSLT14 De-En translation task. We see that the BLEU of the basic NAT model could be enhanced via either of the two regularization terms by about $1$ point. As a comparison, simply de-duplicating the repeated tokens brings certain level of gain, and with the universal regularization that simply penalizes the cosine similarity of all $(h_t, h_{t+1})$, the performance even drops (from $28.73$ to $28.32$). When combining both regularization strategies, the BLEU score goes higher, which shows that the two regularization strategies are somewhat complementary to each other. A noticeable fact is that the gain of combining both regularization strategies (about $2.1$) is lower than the sum of each individual gain (about $2.8$). One possible explanation may be that the two types of errors ({\em repeated translation} and {\em incomplete translation}) can be correlated to some extent and tackling one might help the alleviation of the other. For example, intuitively, given the fixed target side length $T_y$, if repeated tokens are removed, more `valid' tokens will occupy the position, consequently reducing the possibility of incomplete translation. 

Furthermore, we evaluate the effectiveness of alleviating repeated translations with our proposed approach. We count the number of de-duplication (de-dup) operations (for example, there are 2 de-dup operations for ``we \&apos;ll see climate \st{climate} change \st{change}"). The per-sentence de-dup operations in IWSLT14 De-En dev set are 2.3 with NAT-BASE, which has dropped to 0.9 with the introduction of similarity regularization, clearly indicating the effectiveness of the similarity regularization.

\section{Conclusions}

In this paper, we proposed two simple regularization strategies to improve the performance of non-autoregressive machine translation (NAT) models, the similarity regularization and reconstruction regularization, which have been shown to be effective for addressing two major problems of NAT models, i.e., the repeated translation and incomplete translation, respectively, consequently leading to quite strong performance with fast decoding speed. 

While the two regularization strategies were proposed to improve NAT models, they may be generally applicable to other sequence generation models. Exploring such potential would be an interesting direction for future research. For example, we can apply our methods to other sequence generation tasks such as image caption and text summarization, with the hope of successfully deploying the non-autoregressive models into various real-world applications. We also plan to break the upper bound of the autoregressive teacher model and obtain better performance than the autoregressive NMT model, which is possible since there is no gap between training and inference (i.e., the {\em exposure bias} problem for autoregressive sequence generation~\cite{ranzato2015sequence}) in NAT models. 

\bibliographystyle{aaai}
\bibliography{wang}
\end{document}